\documentclass[coverpage]{inftechrep}
\usepackage{mathptmx}
\institutes{Institute for Adaptive and Neural Computation}
\reportnumber{EDI-INF-ANC-1801}
\reportyear{2018}
\reportmonth{9}

\usepackage{subcaption}
\usepackage[utf8]{inputenc} 
\usepackage[T1]{fontenc}    
\usepackage{hyperref}       
\usepackage{url}            
\usepackage{booktabs}       
\usepackage{amsfonts}       
\usepackage{nicefrac}       
\usepackage{microtype}      
\usepackage{graphicx}
\usepackage{amsmath,amssymb}
\usepackage{amsfonts}       
\usepackage{color}
\usepackage{transparent}

\usepackage{amsmath}
\usepackage{enumerate}

\usepackage{nicefrac}

\usepackage{placeins}

\title{GINN: Geometric Illustration of Neural Networks}

\author{
Luke N. Darlow \hspace{2mm} and\hspace{2mm} Amos J. Storkey  \\
School of Informatics\\
University of Edinburgh \\
10 Crichton St, Edinburgh EH8 9AB \\
\texttt{L.N.Darlow@sms.ed.ac.uk \hspace{5mm} a.storkey@ed.ac.uk}  \hspace{9mm}
}

\begin{document}

\maketitle

\begin{abstract}
  This informal technical report details the geometric illustration of decision boundaries for ReLU units in a three layer fully connected neural network. The network is designed and trained to predict pixel intensity from an $(x, y)$ input location. The Geometric Illustration of Neural Networks (GINN) tool was built to visualise and track the points at which ReLU units switch from being active to off (or vice versa) as the network undergoes training. Several phenomenon were observed and are discussed herein. This technical report is a supporting document to the blog post with online demos and is available at \url{http://www.bayeswatch.com/2018/09/17/GINN/}
\end{abstract}

\section{Introduction}
Neural networks are often regarded as difficult to understand or interpret. Visualising their functionality and how they learn is challenging and has driven many research and engineering pursuits. In this technical report, we will take a step back from complex and high-dimensional domains (such as natural images). Instead, we structure a fairly simply problem that is neatly visual in order to interrogate the properties of neural networks.

\begin{figure}[!htbp]
\begin{center}
{\includegraphics[width=0.8\linewidth]{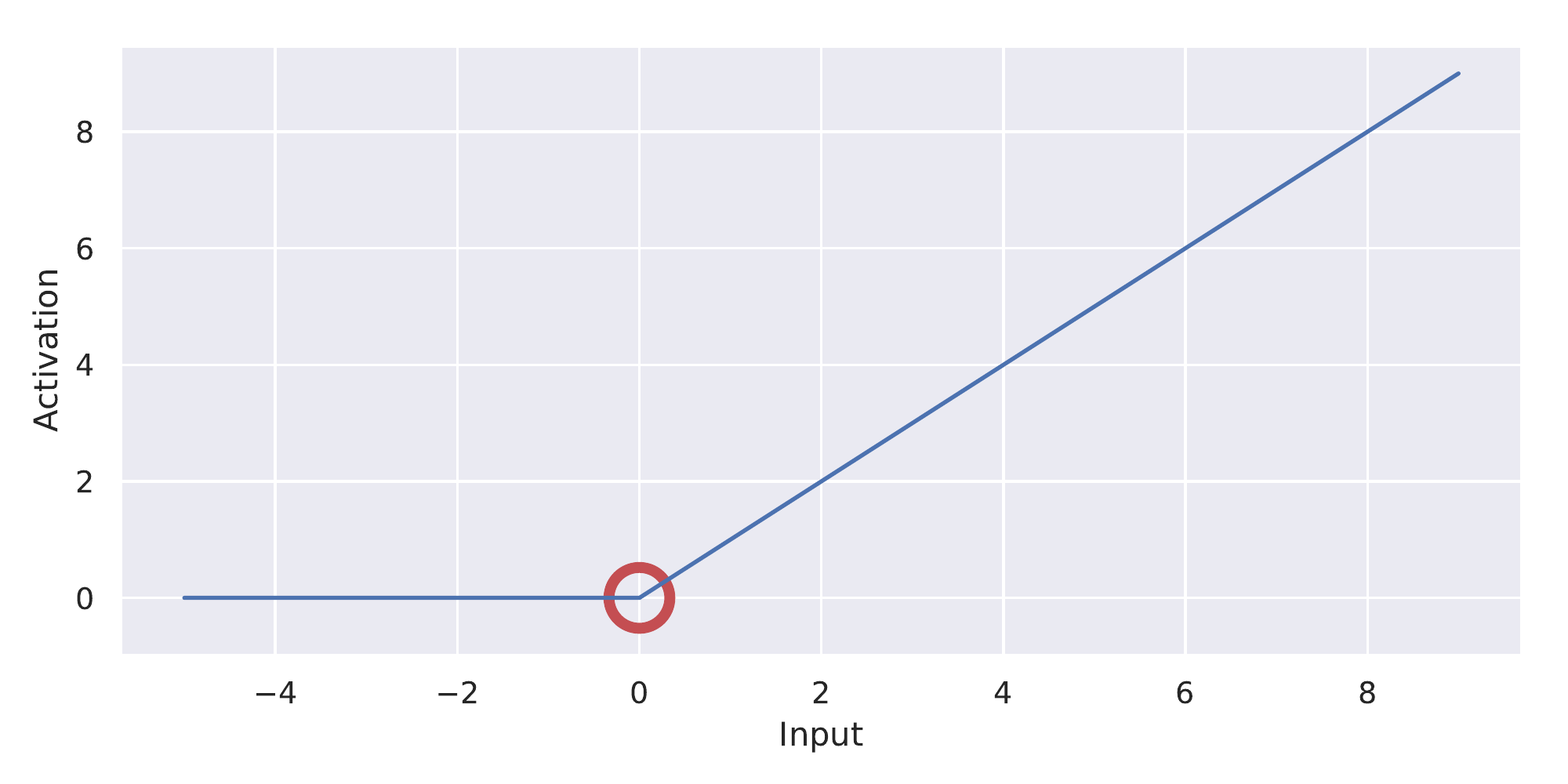}}
\end{center}
  \caption[ReLU activation function]{ReLU activation function: $\mathrm{max}(0, x)$. The red circle indicates this non-linearity decision boundary that we visualise in this technical report.}
\label{fig:relu}
\end{figure}

It is typically not useful to consider the elementary components (neurons) that constitute neural networks. From layer to layer, it is the weighted linear aggregation of inputs and adjustments (biases) that are processed by a non-linear activation function. This particular combination of inputs dictates how a neuron makes use of its activation function. For the purposes of this report, we will only consider the ReLU (Figure \ref{fig:relu}).

The non-linearity boundary of the ReLU is marked with the red circle in Figure \ref{fig:relu}; we call this the \emph{reflex} point. This is the point at which the incoming pre-activation either passes through (if it is positive) or is replaced with zero (if it is negative).  Despite being a simple function -- the maximum of zero or the input -- the ReLU class of non-linearities is widely used. 

It is usually not the activation function itself that allows a neural network to approximate complex functions, but rather the global interaction between all the processing elements. However because the ReLU is piecewise linear, the composition of many ReLU nonlinearities in any layer of a neural network makes the whole layer piecewise linear. This opens up a particular form of visualisation: it is the parts of space that correspond to these ReLU boundaries that dominate the computational power of a neural network.

In this report we will study both the interplay between neurons at different layers as well as single neurons themselves. We do this by inspecting the boundary associated with the reflex point of each neuron \textbf{over an entire data-domain}. A simplified and illustrative data domain is necessary for this (Section \ref{sec:setup}). The Geometric Illustration of Neural Networks (GINN) tool was built to enable exploration of these boundary visualisations through training.

\section{Background}
Since feedforward neural networks came into play in the mid 1980s, there have been many studies analysing and visualising such networks. Typically this has involved visualisation of decision boundaries, generalising from the linear boundaries of simple linear classifiers or perceptrons. Analysis of the benefits of multiple layers to the flexibility of decision boundaries played a part in a number of studies (e.g. \cite{lippman87}), and was the focus of a number of general approximation proofs \cite{cybenko89, funahashi89, hornik89}. These ideas were explored in key neural network texts \cite{bishop_nn_for_pattern_recognition}. However the early adaptation of the sigmoid activation function hindered the visualisation of neural networks somewhat: the complex smooth non-linear surfaces given by sigmoidal neural networks even with few layers were very hard to decompose. Even where decision boundaries were illustrated, tracking these across the learning process was not done: there was more focus on the representation power of the neural network than a desire to follow the learning process. Nowadays, as people consider the unreasonable effectiveness of stochastic gradient methods (e.g. \cite{jastrzebski_threefactors, jastrzebski_sharpest}), visualising learning processes can be insightful.

Because of the difficulty of visualising the computation of a neural network, it became more common to visualise the action of a neural network on a given example: the activations at different layers can be represented pictorally in terms of their response. Presentations of activations were common across many papers, especially in the setting of convolutional networks \cite{lecun1998gradient}.

For large scale convolutional neural networks, visualising the \emph{computation} of the neural network in a unified diagram is still a substantial challenge. In this informal technical report and associated blog\footnote{http://www.bayeswatch.com/2018/09/17/GINN/}, we instead focus on looking at simple scenarios of two dimensional images; in these settings it turns out we can visualise the network's computation if we use rectified linear activations, and we argue these visualisations give us insights into neural network learning and computation.

\section{The Set-up} \label{sec:setup}
To visualise the interaction between an entire data domain and a neural network, we consider the task of predicting pixel intensity from a pixel location (i.e., a two-dimensional input). \textbf{The input is not the entire image but rather individual $(x, y)$ pixel locations (normalised between [0, 1] for simplicity)}. The predicted log-probability output of the neural network answers: is this pixel black or is this pixel white? We set this up as a typical classification task.

\begin{figure}[!htbp]
\begin{center}
{\includegraphics[width=0.9\linewidth]{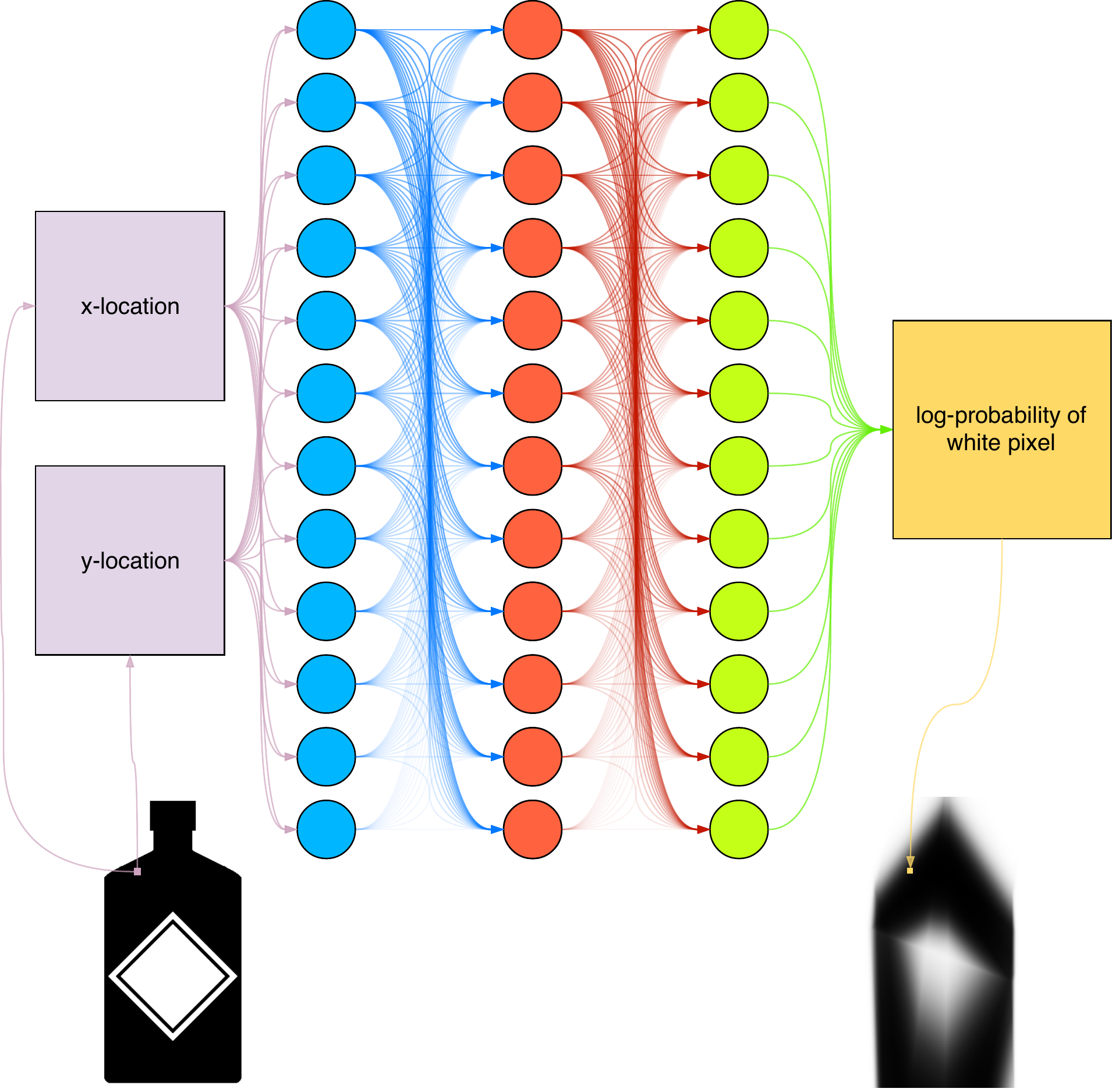}}
\end{center}
  \caption{Network Diagram. Each arrow represents a learnable weight. The \textbf{entire target data domain} is the bottle image on the left, while the image on the right is the corresponding prediction (exemplified here as early stage training) made by the network when processing pixel locations over this domain.}
\label{fig:network}
\end{figure}

The network constructed for this illustration is shown in Figure \ref{fig:network}. It has three hidden layers, each of which have sixteen neurons. It was trained by minimising the cross-entropy loss over 1,280,000 iterations, each of which consists of a randomly selected mini-batches of 128 pixel locations. We used the Adam optimiser with its default hyper-parameters, and gradually reduce the learning rate to zero using cosine annealing. An implementation of this network and the corresponding visualisation tool is available here: \url{https://github.com/learning-luke/ginn}.

The \textbf{entire target data-domain is the image on the left}. Therefore, the function that this neural network is modelling can be visualised and interpreted by processing all pixel locations to produce the image on the right, for example.

\section{The Illustration}
The illustration demonstrates how the neurons in this network behave throughout learning. We built an accompanying web-based GINN tool that can be found here: \url{http://www.bayeswatch.com/assets/ginn/ginn.html}. We completed three `good’ training runs of the network that converged satisfactorily and one `bad’ run that failed to model the central diamond of the input image (see the lower left image in Figure \ref{fig:network}). The truly instructive component of this report involves using this tool. We therefore encourage the reader to do so. Analysing a single neuron’s non-linear boundary throughout training gives an intuition for how these elements coalesce into powerful function approximators.

\begin{figure}[!htbp]
\centering
\begin{subfigure}[b]{0.31\textwidth}
  \centering
  \includegraphics[width=\textwidth]{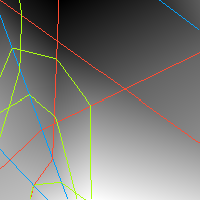}
  \caption{Initialisation}
\end{subfigure}
\begin{subfigure}[b]{0.31\textwidth}
  \centering
  \includegraphics[width=\textwidth]{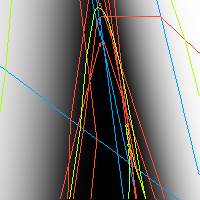}
  \caption{Fitting begins}
\end{subfigure}
\begin{subfigure}[b]{0.31\textwidth}
  \centering
  \includegraphics[width=\textwidth]{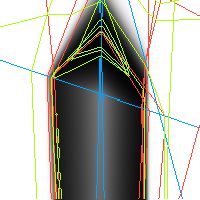}
  \caption{Diamond fitting begins}
\end{subfigure}\\
\begin{subfigure}[b]{0.31\textwidth}
  \centering
  \includegraphics[width=\textwidth]{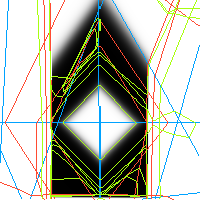}
  \caption{General shape found}
\end{subfigure}
\begin{subfigure}[b]{0.31\textwidth}
  \centering
  \includegraphics[width=\textwidth]{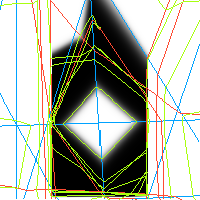}
  \caption{Fine detail fitting begins}
\end{subfigure}
\begin{subfigure}[b]{0.31\textwidth}
  \centering
  \includegraphics[width=\textwidth]{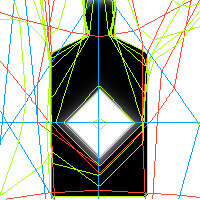}
  \caption{End result}
\end{subfigure}
\caption{A full run of neural network training. The background intensity is the log-probability that the pixel is white, as predicted by the network when processing \textbf{all of the input data}. The blue lines are drawn by detecting where the first hidden layer's neurons switch on or off, while the red and green lines correspond to the second and third hidden layers, respectively.}\label{fig:sweep}
\end{figure}

The visualisation component involves tracking the non-linear activation boundary defined by the ReLU function: when it moves from a region in \textbf{data domain} that triggers the activation to be on or off. Stacking these non-linear boundaries can be interpreted as inducing a more finely granulated interpretation of the data domain from which to make decisions. Specifically, this means that later layers can learn to branch off earlier layers and stacking this behaviour enables a finer division of the data domain (see Figure \ref{fig:sweep} for an example).

It can be difficult to understand what neural networks are learning, how they arrive at a converged state and solution, or where they put their `energy’ (e.g. decision making power, in this case). What GINN offers is an attempt to give an intuitive grasp as to how layers in a neural network interpret the data domain.

Figure \ref{fig:sweep} shows a full run from initialisation to convergence. Each line corresponds to a non-linear boundary of a ReLU unit. While the units in the first hidden layer (blue) can only ever divide the data domain (the background image) into linear regions, the units in the second hidden layer (red) can deviate across the first layer's boundaries -- Figure \ref{fig:sweep} (a) shows this clearly. Furthermore, the third hidden layer boundaries can also be redirected due to the changes in values of the second hidden layer. 

The following sections (\ref{sec:bias} through \ref{sec:failure}) discuss several intriguing phenomena that were noticed when exploring the network's training dynamics using the GINN tool.

\section{Bias Shifts Before Weights Shift}\label{sec:bias}
Changing the bias term for any given neuron translates into shifting a line without changing its slope, while a change of incoming weights alters the slop of a line. Figure \ref{sec:bias} visualises the early stages of learning and reveals that the bias terms are readily and quickly adjusted before weight changes play any significant role (for layer 1, here).

\begin{figure}[htbp]
\centering
\begin{subfigure}[b]{0.31\textwidth}
  \centering
  \includegraphics[width=\textwidth]{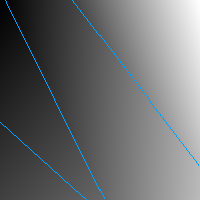}
  \caption{Initialisation}
\end{subfigure}
\begin{subfigure}[b]{0.31\textwidth}
  \centering
  \includegraphics[width=\textwidth]{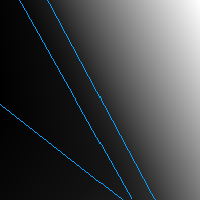}
  \caption{Bias shift only}
\end{subfigure}\\
\begin{subfigure}[b]{0.31\textwidth}
  \centering
  \includegraphics[width=\textwidth]{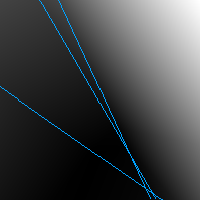}
  \caption{Bias shift and weight shift}
\end{subfigure}
\begin{subfigure}[b]{0.31\textwidth}
  \centering
  \includegraphics[width=\textwidth]{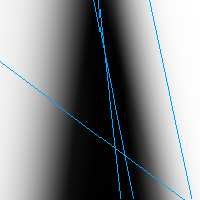}
  \caption{Weight shift (slope changes)}
\end{subfigure}
\caption{The bias shift in the early stages (a) of learning is much greater than any weight changes (d). A shift in bias is illustrated by a parallel shift in location of the lines and weight changes alter the slope of the lines.}\label{fig:sweep2}
\end{figure}

At early training the bias changes dominate, and thereafter the weights are adjusted to change the slope of the lines. This is also true for deeper layers and even seems to happen at multiple instances during training.

\section{Convergence to Critical Points}\label{sec:critical}

\begin{figure}[!htbp]
\centering
\begin{subfigure}[b]{0.31\textwidth}
  \centering
  \includegraphics[width=\textwidth]{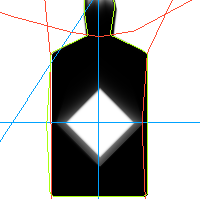}
  \caption{}
\end{subfigure}
\begin{subfigure}[b]{0.31\textwidth}
  \centering
  \includegraphics[width=\textwidth]{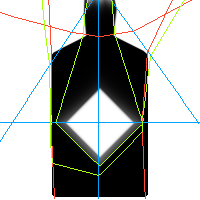}
  \caption{}
\end{subfigure}
\caption{Two examples of how the network's non-linear boundaries tend to converge on critical points in the data domain. }\label{fig:critical}
\end{figure}

Consider Figure \ref{fig:critical}. First, note how the latest layer (green, layer 3) can deviate in response to the boundaries of both earlier layers even though it is not connected to the earliest layer (blue, layer 1). Second, note how the critical points in the data (i.e corners of the bottle) dictate the position of the boundaries in early as well as later layers.

\section{Copycat Neurons}\label{sec:copycats}

\begin{figure}[!htbp]
\centering
\begin{subfigure}[b]{0.31\textwidth}
  \centering
  \includegraphics[width=\textwidth]{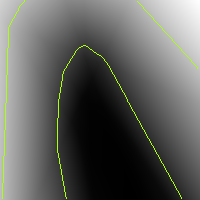}
  \caption{Initialisation}
\end{subfigure}
\begin{subfigure}[b]{0.31\textwidth}
  \centering
  \includegraphics[width=\textwidth]{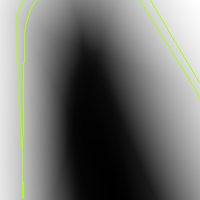}
  \caption{Early training}
\end{subfigure}\\
\begin{subfigure}[b]{0.31\textwidth}
  \centering
  \includegraphics[width=\textwidth]{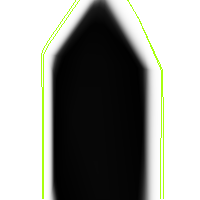}
  \caption{Later training}
\end{subfigure}
\begin{subfigure}[b]{0.31\textwidth}
  \centering
  \includegraphics[width=\textwidth]{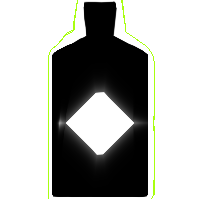}
  \caption{Convergence}
\end{subfigure}
\caption{Copycat neurons. Even though they are initialised differently (a), and they are not identical during training (b) and (c), they end up at a very similar point at convergence (d). }\label{fig:copycat}
\end{figure}

Different neurons sometimes take on identical roles. This is obviously inefficient, but sometimes difficult to avoid. In the case shown in Figure \ref{fig:copycat}, two neurons in the final layer converge to a similar state.

\section{The Indecisive Neuron}
\begin{figure}[!htbp]
\centering
\begin{subfigure}[b]{0.31\textwidth}
  \centering
  \includegraphics[width=\textwidth]{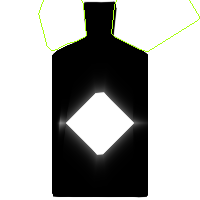}
  \caption{State 1}
\end{subfigure}
\begin{subfigure}[b]{0.31\textwidth}
  \centering
  \includegraphics[width=\textwidth]{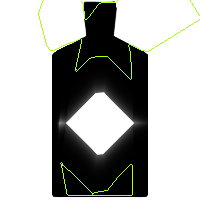}
  \caption{State 2}
\end{subfigure}
\caption{An example of an indecisive neuron. This is the same neuron's decision boundary. At late stage training it flips between these two states without causing a change in loss. }\label{fig:indecisive}
\end{figure}

Sometimes neurons converge to a state where small shifts in the loss cause large changes in their decision boundaries. Figure \ref{fig:indecisive} is an illustration constructed toward the end of a training run, where the learning rate was close to zero (since it was cosine annealed). This behaviour does not seem to change the predicted output, nor is it obvious how this observation could be useful, but it is interesting to witness, nonetheless.

\section{Leveraging Symmetries in the Data}\label{sec:symmetries}

\begin{figure}[!htbp]
\begin{center}
{\includegraphics[width=0.9\linewidth]{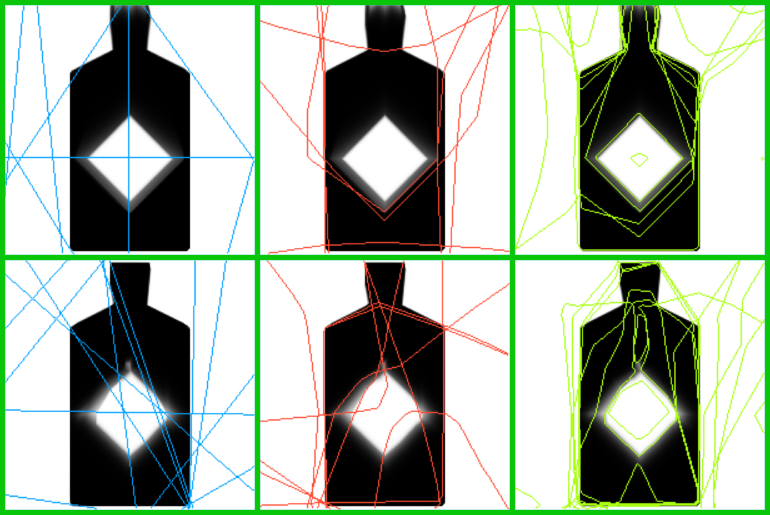}}
\end{center}
  \caption{An example of how the network could either learn to leverage the symmetries in the data (top row) or not (bottom row). When it does, the approximation is usually better.}
\label{fig:symmetry}
\end{figure}

When the network leverages the symmetries in the data, the output prediction becomes better: individual neurons can multitask. Figure \ref{fig:symmetry} demonstrates the difference between two converged networks. The better network (top row) does a better job at predicting the data and the non-linear boundaries in all layers exhibit some left-right symmetry inherent in the data. The worse network (bottom row) does not capture the data symmetry nearly as well.

\section{Failure Case}\label{sec:failure}

\begin{figure}[!htbp]
\begin{center}
{\includegraphics[width=0.32\linewidth]{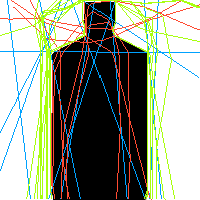}}
\end{center}
  \caption{An example of a failure case.}
\label{fig:failure}
\end{figure}

Figure \ref{fig:failure} demonstrates the converged state of a failure case. It is unclear precisely what goes wrong to result in this poor reconstruction. However, it might be a result of the fact that earlier layers do not find the natural symmetry of the data, or that the initialisation results in later layers favouring certain neurons in earlier layers, thus bounding them within regions that are not useful for modelling the central diamond shape. 

\section{Conclusion}
We argue that the solutions that neural networks find to even simple problems are non-trivial to understand, and good visualisation can help in understanding those solutions. Even more important, the learning process of SGD can be illustrated, and such lessons may help us in understanding how learning of neural networks can be improved. This report outlines some intriguing lessons that can be learned when exploring even the simplest of problems: the decision boundaries of ReLU units in a three layer fully-connected neural network, designed to approximate an image as a function of pixel locations. We hope that these insights may help researchers in understanding how to improve our current choices of optimisation method, initialisation and model form. More importantly we hope that researchers can use this tool to obtain further insights that might further aid understanding of neural network computation.
\bibliographystyle{unsrt}
\bibliography{mybib}

\begin{thebibliography}{1}

\bibitem{lippman87}
R.~Lippmann.
\newblock An introduction to computing with neural networks.
\newblock {\em IEEE ASSP Magazine}, pages 4--22, 1987.

\bibitem{cybenko89}
G.~Cybenko.
\newblock Approximations by superpositions of a sigmoidal function.
\newblock {\em Ath. Control Signals Systems}, 2:303--314, 1989.

\bibitem{funahashi89}
K.~Funahashi.
\newblock On the approximate realization of continuous mappings by neural
  networks.
\newblock {\em Neural Networks}, 2:183--192, 1989.

\bibitem{hornik89}
K.~Hornik, M.~Stinchcombe, and H.~White.
\newblock Multilayer feedforward netowrks are universal approximators.
\newblock {\em Neural Networks}, 2:359--366, 1989.

\bibitem{bishop_nn_for_pattern_recognition}
C.~M. Bishop.
\newblock {\em Neural Networks for Pattern Recognition}.
\newblock Oxford University Press, 1995.

\bibitem{jastrzebski_threefactors}
S.~{Jastrz{\c e}bski}, Z.~{Kenton}, D.~{Arpit}, N.~{Ballas}, A.~{Fischer},
  Y.~{Bengio}, and A.~{Storkey}.
\newblock Three factors influencing minima in {SGD}.
\newblock In {\em Proceedings of ICANN2018}, 2018.

\bibitem{jastrzebski_sharpest}
S.~{Jastrz{\c e}bski}, Z.~{Kenton}, N.~{Ballas}, A.~{Fischer}, Y.~{Bengio}, and
  A.~{Storkey}.
\newblock {DNN}'s sharpest directions along the {SGD} trajectory.
\newblock {\em ArXiv e-prints}, July 2018.

\bibitem{lecun1998gradient}
Yann LeCun, L{\'e}on Bottou, Yoshua Bengio, and Patrick Haffner.
\newblock Gradient-based learning applied to document recognition.
\newblock {\em Proceedings of the IEEE}, 86(11):2278--2324, 1998.

\end{thebibliography}
\end{document}